\documentclass[conference]{IEEEtran}
\usepackage{cite}
\IEEEoverridecommandlockouts
\usepackage{cite}
\usepackage{amsmath,amssymb,amsfonts}
\usepackage{algorithmic}
\usepackage{graphicx}
\usepackage{textcomp}
\usepackage{xcolor}
\usepackage{booktabs}
\usepackage[colorlinks,linkcolor=blue,urlcolor=black]{hyperref}
\newcommand{\etal}{\textit{et al.}}
\def\BibTeX{{\rm B\kern-.05em{\sc i\kern-.025em b}\kern-.08em
    T\kern-.1667em\lower.7ex\hbox{E}\kern-.125emX}}

\title{Coordinating Cross-modal Distillation for Molecular Property Prediction
\thanks{* These authors contributed equally.\\}
}

\begin{document}

\author{
\IEEEauthorblockN{
Hao Zhang\textsuperscript{2,*}
Nan Zhang\textsuperscript{1,*},
Ruixin zhang\textsuperscript{2},
Lei Shen\textsuperscript{2},
Yingyi Zhang\textsuperscript{2},
Meng Liu\textsuperscript{3}
}
\IEEEauthorblockA{
\textsuperscript{1}\textit{
Academy for Engineering and Technology,
Fudan University, China}\\
\textsuperscript{2}\textit{
Tencent Youtu Lab}\\
\textsuperscript{3}\textit{
Shanghai Jiao Tong University}\\
}
}

\maketitle

\begin{abstract}

   In recent years, molecular graph representation learning (GRL) has drawn much more attention in molecular property prediction (MPP) problems. The existing graph methods have demonstrated that 3D geometric information is significant for better performance in MPP. However, accurate 3D structures are often costly and time-consuming to obtain, limiting the large-scale application of GRL. It is an intuitive solution to train with 3D to 2D knowledge distillation and predict with only 2D inputs. But some challenging problems remain open for 3D to 2D distillation. One is that the 3D view is quite distinct from the 2D view, and the other is that the gradient magnitudes of atoms in distillation are discrepant and unstable due to the variable molecular size. To address these challenging problems, we exclusively propose a distillation framework that contains global molecular distillation and local atom distillation. We also provide a theoretical insight to justify how to coordinate atom and molecular information, which tackles the drawback of variable molecular size for atom information distillation. Experimental results on two popular molecular datasets demonstrate that our proposed model achieves superior performance over other methods. Specifically, on the largest MPP dataset PCQM4Mv2 served as an "ImageNet Large Scale Visual Recognition Challenge" in the field of graph ML, the proposed method achieved a 6.9\% improvement compared with the best works. And we obtained fourth place with the MAE of 0.0734 on the test-challenge set for OGB-LSC 2022 Graph Regression Task. We will release the code soon.

\end{abstract}

\section{Introduction}
In recent years, molecular property prediction (MPP) has played a crucial role in new materials and drug discovery. 
Since atoms and bonds in a molecule are naturally in a graph structure, graph neural networks (GNN) \cite{scarselli2008graph} have shown great vitality in MPP. 
Many works \cite{gilmer2017neural,liu2019n,xu2021self} take 2D molecular structures as input of GNN for training and testing. 
Their performances are limited because they do not consider the 3D atom coordinates that determine certain chemical and physical functionalities of molecules. 
Methods \cite{klicpera2020directional,qiao2020orbnet,liu2021spherical} that exploit 3D structures exhibit advantages in accuracy, especially properties related to quantum mechanics, such as single-point energy, atomic forces, and dipole moments.
However, obtaining accurate 3D structures is expensive and time-consuming.
Calculating 3D information with DFT takes ${10^5}$ times longer than model feedforward \cite{gilmer2017neural}, which seriously impedes the application of 3D-based methods. 

Some works attempt to augment 2D information with 3D information through training to enhance the prediction accuracy while keeping inference speed.
One direction is developing 2D-3D contrastive tasks. Another way is to generate original conformations through reconstruction tasks. 
Usually, these tasks require several different 3D conformers to create positive and negative pairs for self-supervised learning, but obtaining different conformers is time-consuming for large-scale data. 
Also, different conformers from one SMILES\cite{toropov2005simplified} code may represent different property values, which lower the performance of contrastive-based methods. Lastly, some applications do not support the requirement of multi conformers.
With the force field given, one SMILES code usually can only generate one 3D conformer structure.

Distillation is another intuitive solution that utilizes 3D knowledge in training while keeping inference speed. 
Also, distillation methods do not have the limitation of multiple conformers requirements. 
But some challenging problems remain for distillation methods. On the one hand, the 3D view is quite distinct from the 2D view, making the top-level-feature-only distillation less effective. 
On the other hand, which tokens participate in distillation is still an open question for the MPP. 
As self-attention calculates information of the virtual token from all-atom features, it is natural to include all-atom tokens together in distillation.
But simply distilling atom tokens and virtual token together leads to poor performance.
We found the gradient magnitudes of atom tokens distillation are inconsistent and unstable due to the variable molecular size. 
So distilling local atom information without correction will cause degradation problems. 

Motivated by the above observation, we propose a novel coordinating cross-modal distillation (CCMD) framework. 
We formulate a coordinating weight, which dynamically scales and balances the gradients of global molecular token and local atom tokens according to the number of atoms. 
Also, the distillation of virtual token and atom tokens is carried in all layers to enhance the 2D-3D distillation performance. 
Lastly, we introduce absolute position encoding to integrate adjacent edge information, achieving better performance in both views of 2D and 3D.
Experimental results on two popular molecular datasets demonstrate that our proposed model achieves state-of-the-art performance.

In conclusion, our main contributions to this work can be summarized as follows:
\begin{itemize}
\item We exclusively develop an enhanced cross-modal distillation on all layers for molecular property prediction, which distills global molecular information and local atom information to transfer 3D geometric information to 2D view.
\item 
For coordinating global molecular and local atom information, we formulate a novel weight according to the number of atoms in a molecular to scale gradient magnitudes, which boosts the distillation performance.
Specific, we demonstrate the coordinating weight is $f(N)=\frac{1}{{N}^2}$ for transformer and $f(N)=\frac{1}{{N}}$ for GIN. N is the number of atoms in a molecular.
\item We design an absolute position encoding integrated with adjacent edge information, which could distinguish atoms same atom id in the input layer and increase the performance of both 2D and 3D.
\item CCMD can enable molecular attention to more atoms in the 2D view. Extensive experiments on the PCQM4Mv2 dataset \cite{hu2021ogb} demonstrate that our proposed method consistently outperforms other state-of-the-art methods in most situations.
\end{itemize}

\section{Related Work}
\subsection{Molecular Property Prediction}
According to different dimensions, molecular expressions contain 1D SMILES code, 2D graph, and 3D graph. 
Due to the natural capability for representing the molecular structure, graph neural networks (GNN) have attracted attention in molecular property prediction, especially with the input of 2D or 3D expressions. 
\subsubsection{2D Methods}

Many works focus on applying GNN to 2D graphs.
For example, Gilmer \etal \cite{gilmer2017neural} proposed Message Passing Neural Networks, which combine message passing and aggregating algorithms with GNN, forming a successful framework for MPP.
Liu \etal \cite{liu2019n} introduced a simple unsupervised representation for molecules that embeds the vertices in the molecule graph and constructs a compact feature by assembling the vertex embeddings.
Ying \etal \cite{ying2021transformers} proposed Graphormer, which utilizes a Transformer \cite{vaswani2017attention} in the GNN by effectively encoding the graph structure information. 
Although these methods have shown promising capabilities, their potential is limited as they do not use 3D coordinates that better represent molecular structure and energy.

\subsubsection{3D Methods}
3D molecular structures contain spatial information important to molecular property prediction, such as bond angles and lengths. 
Many recent methods try to answer how to utilize this 3D information in the GNN model. 
For instance, Klicpera \etal \cite{klicpera2020directional} proposed to let message embeddings interact based on the distance between atoms and the angle between directions to improve quantum mechanical property prediction. 
Lu \etal \cite{lu2019molecular} proposed a Multilevel Graph Convolutional neural Network that extracts features from the conformation and spatial information with multilevel interactions. 
Liu \etal \cite{liu2021spherical} proposed SphereNet, a 3D graph network framework with spherical message passing.
3D-based methods achieve significant performance improvements. But obtaining 3D structures is time-costing, which reduces the value of 3D-based methods in large-scale applications. 

\subsection{Cross Modality Training Methods}
Since training is relatively insensitive to time-consuming, an intuitive approach is to use 2D and 3D information in training and only use 2D in prediction. 
\subsubsection{Contrastive Methods}
These methods use 3D conformers and 2D structures to construct positive and negative pairs while applying a contrastive loss during training.
For example, Liu \etal \cite{liu2021pre} proposed the Graph Multi-View Pretraining framework, in which they perform self-supervised learning by leveraging the correspondence and consistency between 2D topological structures and 3D geometric views. Li \etal \cite{li2022geomgcl} developed GeomGCL, a dual-view geometric message passing network that utilizes the molecular geometry across 2D and 3D views. 
Hu \etal \cite{hu2019strategies} proposed a strategy of pre-training GNN at the level of individual nodes as well as entire graph, so the GNN can learn useful local and global information. Stark \etal \cite{stark20223d} proposed 3D Infomax, which pre-trains a GNN by maximizing the mutual information between its 2D and 3D embeddings.
The drawbacks of contrastive methods lie in the requirements of multiple conformers. Large-scale conformer generation is costly, while one SMILES code usually generates only one 3D conformer with a specific force field.

\subsubsection{Knowledge Distilling Methods}
Knowledge distilling (KD) is an intuitive solution for 3D to 2D cross-modality training.  
As our work relies on Transformer-based methods such as Graphormer, we only discuss Transformer related KD methods. 
Many works \cite{jiao2019tinybert,sun2019patient,wang2020minilm,tian2019contrastive,dong2020distilling} have attempted to apply KD to the Transformer-based model.
Upon molecular property prediction, Zhu \etal \cite{zhu2021stepping} proposed ST-KD, an end-to-end Transformer KD framework, bridging the knowledge transfer between graph-based and SMILES-based models. 
But ST-KD only KD the atom to atom attentions, neglecting the informative embeddings in atom tokens. 
Based on our experiment, gradient magnitudes of atom-token distillation are discrepant and unstable due to the variable molecular size. Distilling atom-token embeddings without correction will cause degradation problems.

To address this issue, we propose a theoretical insight to justify how to coordinate atom token embeddings and molecular features and formulate an auto-correcting distilling weight to dynamically tune and balance the gradients according to the number of atoms.

\section{Method}
\begin{figure*}[t]
  \centering
  \includegraphics[width=\linewidth]{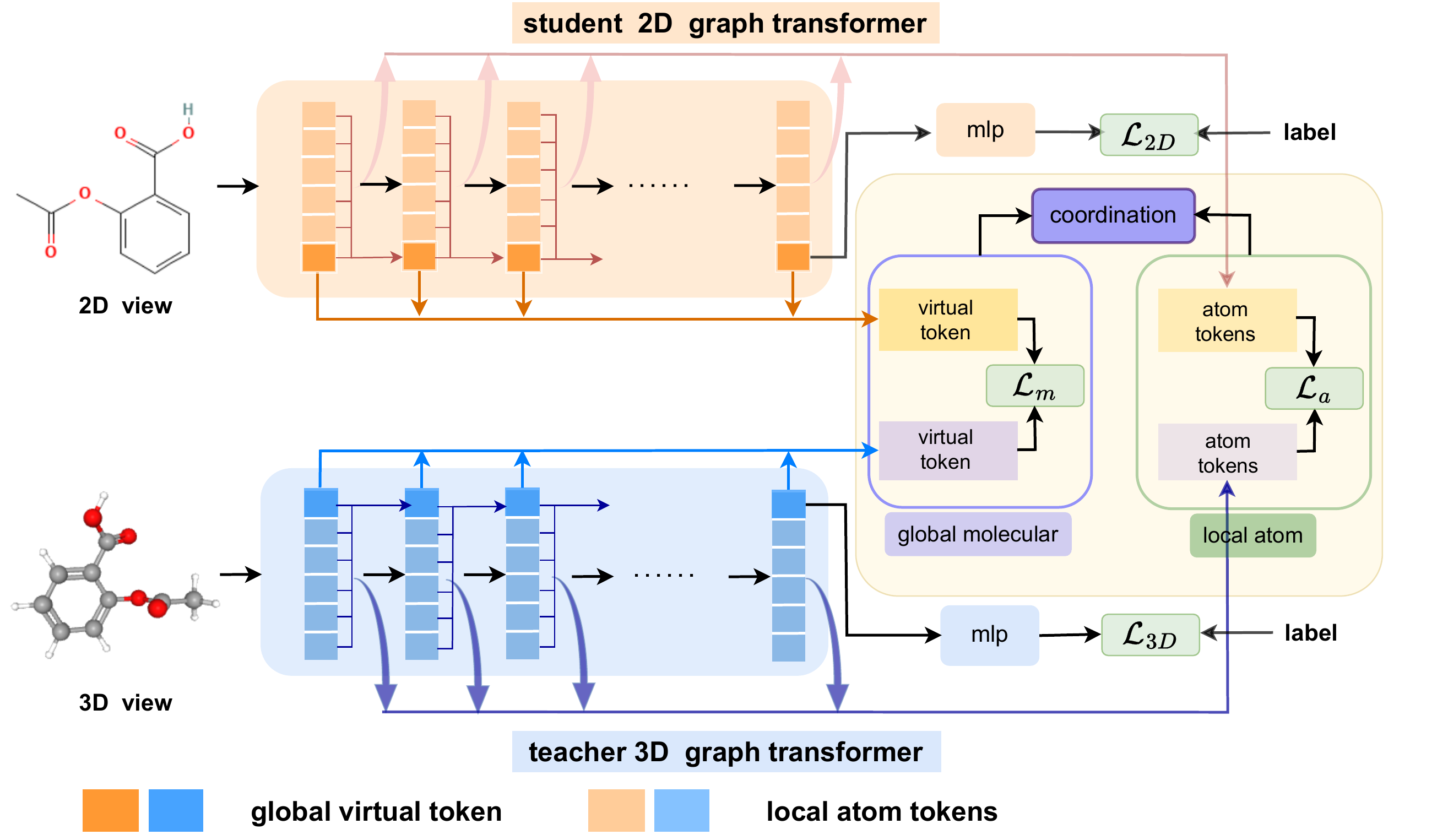}
  \caption{The illustration of of our CCMD framework with global molecular and local atom information for molecular property prediction.
  $\mathcal{L}_{m}$ represents the global molecular distillation, $\mathcal{L}_{a}$ represents the local atom information distillation, and coordinating denotes the coordinating weight scaling gradient magnitudes.
}
  \label{fig:framework}
\end{figure*}

\subsection{Overview}
In this section, we present the key components of CCMD. As illustrated in Fig. \ref{fig:framework}, the whole graph transformer framework consists of a 3D model and a 2D model, and the backbone is based on the graphormer \cite{ying2021transformers}. 
Firstly, the 3D model is trained with 3D information as a teacher model. 
Then, frozing the parameters of the teacher, we train the 2D student model with a distilling loss and a supervised target loss. The distilling module consists of the global molecular distillation and local atom distillation, which can further improve the performance of 2D molecular graphs. At last, for coordinating global and local information, we formulate a novel coordinating weight to scale and balance the gradients optimally.

 \subsection{Preliminary}
 
 For convenience, let $\mathcal{D}=\{(X,Y,E)\}$ denotes a 2D molecular graph, the $X$, $Y$ and $E$ denote the input of atoms, property labels and edges respectively. In addition, we use $\widetilde{}$ mark the 3D view, so 3D molecular graph can be denoted as $ \widetilde{\mathcal{D}}=\{(\widetilde{X},\widetilde{Y},\widetilde{E})\}$, where $\widetilde{E}=f(\widetilde{P})$, and $\widetilde{P}$ represents the 3D coordinates of atoms. For instance, the $i$-th atom embedding in $l$-th layer can be denoted as $X_{i}^{l}$, where $i\in [0,N+1]$. $i=0$ denotes the virtual token representing the molecular representation, $N$ denotes the total number of atoms. $l\in [0,L]$ denotes the order number of the layer.




 \subsection{Input Representations}
 \subsubsection{Absolute Position Encoding}
 Following the previous works (Park et al. 2022; Ying et al.
2021), we integrate an absolute position encoding (APE) into graphormer \cite{ying2021transformers} illustrated in Fig.2. The APE is the adjacent edge information of the node, which could distinguish some atoms with the same atom id easily in
the input layer. Let $X^{0}$ denote the input tokens which are integrates with the APE, $X^{0}$ is calculated as:
\begin{equation}
X^{0}=X\oplus E,
\end{equation}
where $\oplus$ is used to gather the information from neighbors.
In specific, the $i$-th input token is calculated as:
\begin{equation}
X_{i}^{0}=MLP(\sum_{i=1}^{N+1}e_{i,j}).
\end{equation}

For the 2D view, ${e}_{i,j}$ is the feature of original Chemical bonds.
For the 3D view, we utilize radial basis functions (RBF) \cite{buhmann2000radial} as the transformed function   
 $\widetilde{e}_{i,j}=RBF(d_{ij})$ to obtain the edge information, where ${d}_{ij}$ is the distance between atom i and atom j. This 3D embedding can encode diverse geometric factors 

\subsection{Backbone}
The backbone used is graphormer \cite{ying2021transformers}. All tokens in the $l+1$ layer is updated by layer normalization (LN) and multi-head attention (MHA) on the tokens in the $l$ layer
\begin{equation}
X^{'(l+1)}=MHA(LN(X^{l}))+X^{l},
\end{equation}
then the feed-forward blocks (FFN) is applied to the tokens in the $l+1$ layer:
\begin{equation}
X^{l+1}=FFN(LN(X^{l+1}))+X^{'(l+1)}.
\end{equation}

Moreover, the final molecular representation $R$ is obtained from the virtual token of the last layer by an MLP:
\begin{equation}
R=MLP(X_{0}^{L}).
\end{equation}

At last, for the supervised task, we use $L1$ loss.  
So the supervised term $\mathcal{L}_{3D}$ is defined as:
\begin{equation}
\mathcal{L}_{3D}(\widetilde{R},\widetilde{Y})=L_{1}Loss(\widetilde{R},\widetilde{Y}),
\end{equation}
The supervised term $\mathcal{L}_{2D}$ is defined as:
\begin{equation}
\mathcal{L}_{2D}(R,Y)=L_{1}Loss({R},{Y}),
\end{equation}


\subsection{Distillation}
\subsubsection{Challenges}
Usually, for most of the existing distillation methods, the total loss is defined as:
\begin{equation}
\mathcal{L}(R,Y,\widetilde{R}) = {L}_{2D}(R,Y) + \mathcal{W} * L_{1}Loss({X},\widetilde{X}),
\end{equation}
${L}_{2D}(R,Y)$ is the task-related loss, and $L_{1}Loss({X},\widetilde{X})$ is the distilling loss on all tokens and ${W}$ is the hyper-parameter. However, the virtual token ${X}_{0}$ is the target, which represents the global representation of the molecular, and the other tokens ${X}_{i}(i>0)$ are the atom representations that are auxiliary to the molecular representation. So the strategy of distillation should be divided into two modules: global
molecular distillation and local atom distillation. In addition, We should select a proper weight to balance the global molecular distillation and local atom distillation.
 \subsubsection{Global molecular distilling}
 The virtual token ${X}_{0}$ represents the molecular embedding and usually be used to predict molecular property, so distilling the virtual token embedding provides a global view of learning on 3D geometry structural knowledge.
 Considering that 2D graphs are seriously distinct from 3D conformers, we conduct the global molecular distilling from 3D to 2D on all layers which could learn more geometric information at different levels step by step. At last, the target is conducted by minimizing the following loss function given any pairs of $\widetilde{X}_{l,0}$ and $X_{l,0}$ in the training batch:
\begin{equation}
 \mathcal{L}_{m} = \sum_{l=1}^{L}  L_{1}Loss(\widetilde{X}_{0}^{l},X_{0}^{l})
\end{equation}
\subsubsection{Local Atom Distilling}

The virtual token ${X}_{0}$ is the global view of molecular, and the token ${X}_{i}(i>0)$ is the local view of molecular. The virtual token is calculated from atom tokens by self-attention with the updating process. Considering the relationship between the virtual token and atom tokens, we use the local atom distilling to further boost the performance of the virtual token.
One possible way is to calculate the similarity (distance) of each pairwise
atom embedding, then the local atom calibration distillation
loss is defined as:
\begin{equation}
 \mathcal{L}_{a} = \sum_{l=1}^{L}\sum_{j=1}^{N+1}  L_{1}Loss(\widetilde{X}_{j}^{l},X_{j}^{l})
\end{equation}

\begin{figure}[!t]
\centerline{\includegraphics[width=\linewidth]{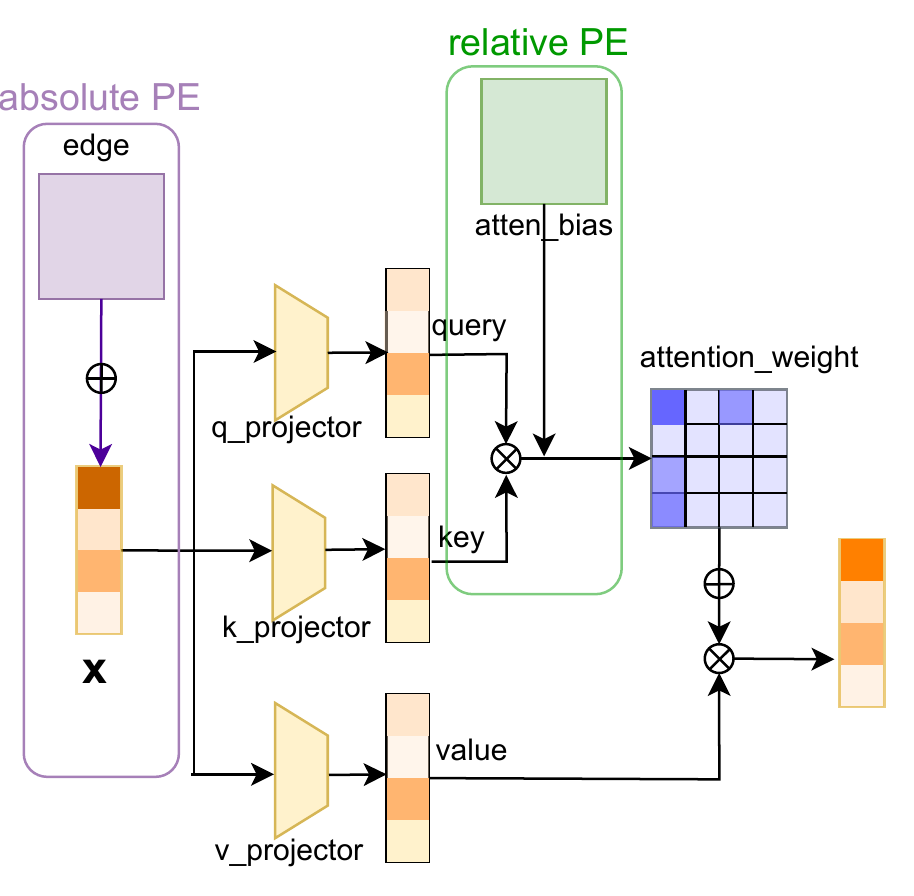}}
\caption{The illustration of the backbone with absolute position encoding and relative position encoding.} 
\label{fig:pe}
\end{figure}
\subsubsection{Coordinating weight}

For boosting the performance of distilling with global and local information, we
formulate a novel coordinating weight according to the number of molecular size to scale gradient magnitudes. The analyzing and reducing process from two perspectives is as follows.

\textbf{Coordinating the total loss} 
Comparing the global molecular loss $\mathcal{L}_{m}$ and local atom loss $\mathcal{L}_{a}$, we can draw the conclusion that $\mathcal{L}_{a}$ is unstable due to the sum operation on all tokens. This means that when the molecular is very large(with big N), the final loss ${L}$ is exploding. So, we should add the function $f(N)$ according to the variable number of atoms to scale and calibrate the local atom loss $\mathcal{L}_{a}$. The formulation is as follows:
\begin{equation}
\begin{aligned}
    \mathcal{L} &=\mathcal{L}_{2D}+(\sum_{l=1}^{L}  L_{1}Loss(\widetilde{X}_{0}^{l},X_{0}^{l}) \\
&+f(N)*\sum_{l=1}^{L}\sum_{j=1}^{N+1}  L_{1}Loss(\widetilde{X}_{j}^{l},X_{j}^{l}),
\end{aligned}
\end{equation}
$L$ is fixed and $N$ is variable which represents the number of atoms. In order to eliminate the effect of sum operation on all tokens, it's intuitive to set $f(N)=\frac{1}{N}$. In addition, we couldn't add $\frac{1}{N}$ to the molecular loss, which will dissipate the loss of the virtual token as N increases, so it's necessary to separate global and local loss, and only scale the atom loss ${{L}_{a}}$. Generally, to make the total loss not explode, we should add
a function $f(N)\propto {O}(\frac{1}{N})$ or $f(N)\propto {o}(\frac{1}{N})$. ${O}$ means “is of the same order as” and ${o}$ means “is ultimately smaller than”.




\textbf{Coordinating the molecular gradient} 
Since the virtual token is the target embedding, its gradient is much important. So we will calculate the relationship between its gradient and N. To simplify the forward of transformer, $X_{i}^{l+1}$ in $(l+1)$ layer is formed as:
\begin{equation}
X_{i}^{l+1}=MHA(X^{l}) \\
\&=\sum_{j=0}^{N+1}({W}^{l}_{ji}*{X}^{l}_{j})
\end{equation}
The ${W}^{l}_{ji}$ is the attention weight. In the back-propagation, the gradient of the virtual token in $l$th layer is as formulated:
\begin{equation}
\begin{aligned}
 {grad(X_{0}^{l})} &= \nabla_{X_{0}^{l}}{L} \\
&=\sum_{k=0}^{N+1}((\nabla_{X_{k}^{l+1}}{L})*(\nabla_{X_{0}^{l}}{X_{k}^{l+1}}))
\end{aligned}
\end{equation}
Finally, for transformer, the virtual token gradient $grad(X_{0}^{l+1})$ in ${l}$ layer can be described in follow:
\begin{equation}
\begin{aligned}
 {grad(X_{0}^{l})} &\propto{{O}({N}^{2})},
 \end{aligned}
\end{equation}
Different from multi-head attention in transformer, a simple GNN always use ${MLP}$ to calculate ${X}^{l+1}_{i}$ with ${X}^{l}_{0}$ . So $(\nabla_{X_{0}^{l}}{X_{k}^{l+1}})$ is constant, then the result is:
\begin{equation}
\begin{aligned}
 {grad(X_{0}^{l})} &\propto{{O}({N})},
 \end{aligned}
\end{equation}
For detailed proof, please refer to the Section1 in Supplementary Material. For transformer, the formula indicates that the gradient of virtual token $grad(X_{0}^{l})$ will be scaled by ${N}^{2}$, and N is the number of atoms. So scaled with $\frac{1}{N}$ couldn't solve the unstable gradient of the virtual token. However, for a simple GNN, $\frac{1}{N}$ is enough. \\
\textbf{Conclusion} 
To this end, for transformer, we proved the weight $f(N)=\frac{1}{{N}^2}$ is suitable to scale the losses and balance the gradient magnitudes of the global molecular information and local atom information. 
So the total objective function is formulated as:
\begin{equation}
\mathcal{L}=\mathcal{L}_{2D}+(\mathcal{L}_{m}+\frac{1}{{N}^{2}}*\mathcal{L}_{a}).
\end{equation}


\section{Experiments}
To evaluate our method and investigate the effectiveness of the proposed components, we carry out extensive experiments on two popular molecular property prediction tasks, especially in PCQM4Mv2 \cite{hu2021ogb}.
In this section, we introduce the detailed configuration in our experiments and present the detailed experimental results.

\subsection{Datasets}
\subsubsection{PCQM4Mv2 dataset} PCQM4Mv2 dataset \cite{hu2021ogb} served as an "ImageNet Large Scale Visual Recognition Challenge" in the field of graph ML is a quantum chemistry dataset originally curated under the PubChemQC project. It contains 3.8 million molecular graphs. The 3D structure for training molecules is calculated by DFT, which is equipped with only one conformer for each SMILES code.

\subsubsection{MolHIV dataset} MolHIV dataset \cite{hu2020open} is adopted from the MOLECULENET \cite{wu2018moleculenet}
with 41K molecular graphs for molecular property prediction.
 All the molecules are
pre-processed using RDKIT \cite{landrum2013rdkit}. Each graph represents a molecule, where nodes are atoms, and edges are chemical bonds. 
Input node features are 9-dimensional, containing atomic number and
chirality, as well as other additional atom features such as formal charge and whether the atom is
in the ring. Input edge features are 3-dimensional, containing bond type, bond stereochemistry as
well as an additional bond feature indicating whether the bond is conjugated.

\subsection{Experimental Setup}

Our experiments are conducted on eight NVIDIA Tesla V100 GPUs.
The implementation is based on PyTorch \cite{2017Automatic}. We use the Graphormer architecture \cite{ying2021transformers} as the based model (L = 12, d = 768) for both 2D and 3D branches. 
Our CCMD method contains improved baseline with APE, global molecular distillation, local atom distillation, searched weight and coordinating weight, denoted as baseline, $\mathcal{L}_{m}$, $\mathcal{L}_{a}$, ${W}_{search}$ and ${W}_{coordinating}$ respectively.
In order to prevent the explosion of a total loss, the local atom distillation loss is calculated by means denoted as $\overline{\mathcal{L}_{a}}$, where $\overline{\mathcal{L}_{a}}=\frac{1}{N}*\mathcal{L}_{a}$.
So we set the coordinating weight as $\frac{1}{N}$ for experiments on graphormer and $1$ for experiments on GIN \cite{xu2018powerful}. 

For the settings and metrics on the PCQM4Mv2 dataset, the previous methods use the 2D view for training and validation. We add the 3D structure calculated by DFT for training and distillation, and also evaluate on the validation with 2D view (the test labels are no longer available and results are given over the validation set). And split of dataset is followed by previous work \cite{ying2021transformers}. We set the batch size as 512 with 70 epochs and use the Adam algorithm \cite{kingma2014adam} with an initialized learning rate of 0.0002 and a momentum of 0.9.  The results are evaluated in terms of Mean Absolute Error (MAE), and a lower MAE indicates better performance.

For the settings and metrics on the MolHIV dataset, we add the 3D structure calculated by RDKIT for 3D view training and distillation. And we evaluate on the test dataset with 2D view. And split of dataset is followed by previous work \cite{ying2021transformers}.
We pretrain 3D and 2D branches on the PCQM4Mv1 dataset following the previous works \cite{ying2021transformers}.
When finetune on the MolHIV dataset, we set the batch size as 64 with 10 epochs and use Adam algorithm \cite{kingma2014adam} with an initialized learning rate of 0.0002 and a momentum of 0.9. The results are evaluated in terms of Area Under the ROC Curve (AUC), and a higher AUC indicates better performance.

\subsection{Experimental Results}

\subsubsection{The comparison on PCQM4Mv2 dataset}
Table \ref{tab:pcqv2} presents the results for molecular property prediction task on the OGB-LSC PCQM4Mv2 datasets \cite{hu2021ogb}. 
What's worth mentioning is that contrastive methods are hard applicable because the PCQM4Mv2 is equipped with only one conformer for each SMILES code. 
Compared with established methods, our model is currently the best on the PCQM4Mv2 leaderboard. As shown in Table \ref{tab:pcqv2}, our approach achieves the significant
MAE improvement by 6\%  compared with the state-of-the-art method, indicating the effectiveness of our proposed method.

\begin{figure}[!t]
\centerline{\includegraphics[width=\linewidth]{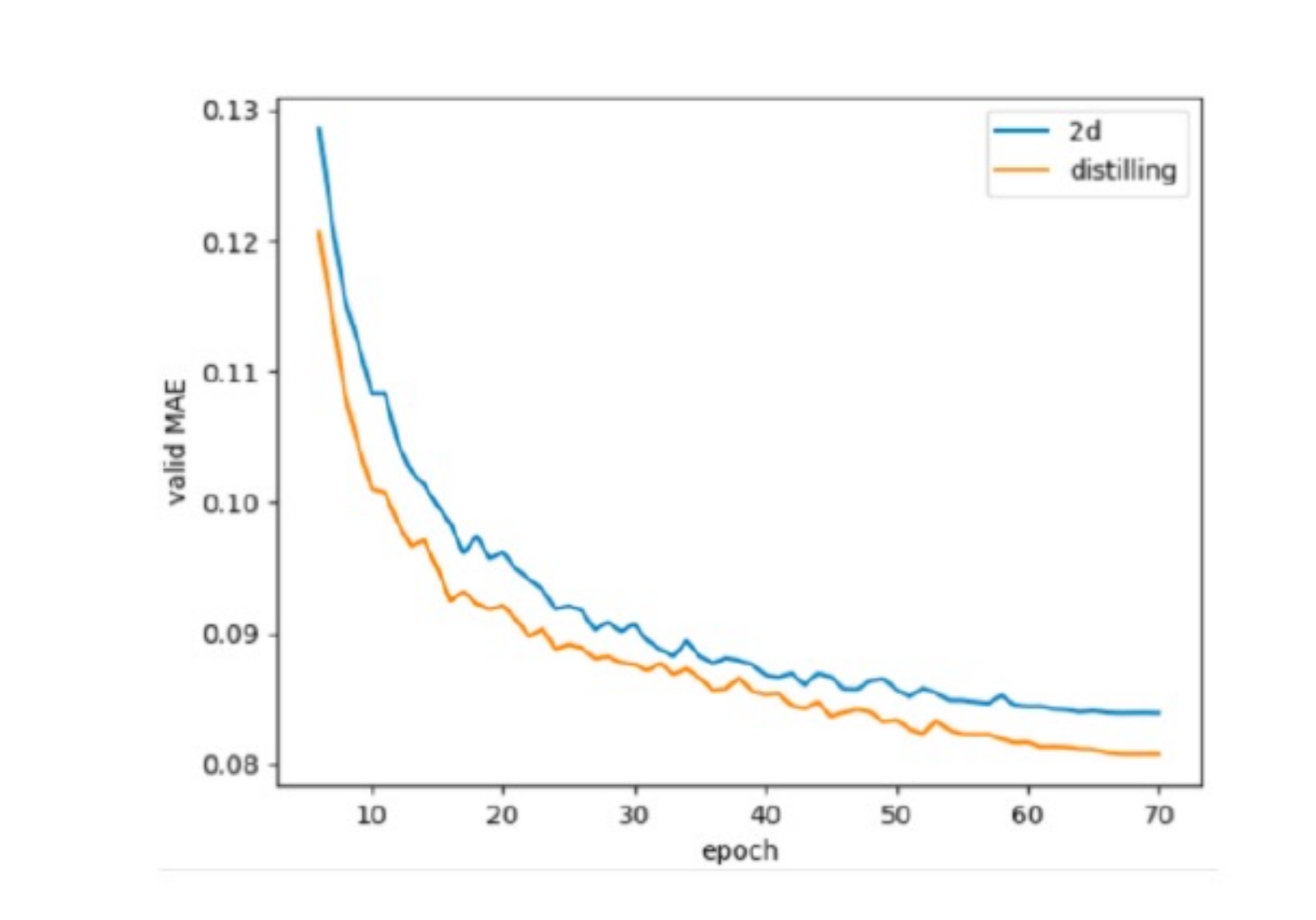}}
\caption{The learning curves of 2D supervised performance and distilling performance during training process.} 
\label{fig:curve}
\end{figure}

\begin{table}[!t]
\caption{The comparison with state-of-the-art methods on the validation set of PCQM4Mv2 dataset. Our approach outperforms the best model on average by 6\%.}
\label{tab:pcqv2}
\centering
\setlength{\tabcolsep}{11.5mm}{
\begin{tabular}{cc}
\toprule
Method & MAE  \\
\midrule
GCN \cite{kipf2016semi}  & 0.1379   \\
GCN-VN\cite{gilmer2017neural,kipf2016semi} & 0.1153  \\
Graphormer  \cite{ying2021transformers} & 0.0864  \\          
EGT  \cite{hussain2022global} & 0.0869  \\
GRPE-Standard  \cite{park2022grpe} & 0.0890  \\
GPS  \cite{rampavsek2022recipe} & 0.0858  \\
\textbf{Ours} & \textbf{0.0809}  \\
\bottomrule
\end{tabular}}
\end{table}

\subsubsection{The comparison on MOLHIV dataset}

We further investigate our proposed method on the MOLHIV dataset. The comparison results are shown in Table \ref{tab:hiv}.
From Table \ref{tab:hiv}, we can see that our method outperforms other methods, achieving prominent improvement by 1.39\% on AUC compared with baseline.
In addition, the origin MOLHIV dataset is not attached with 3D information, so we use the rdkit to generate the 3D position as the 3D coordinates of atoms. 
It should be noted that the generated 3D coordinates with randomly selected 3D conformers are not matched the real coordinates, so the generated force fields are different from the real force fields. The weakness of the 3D teacher model limits the performance of the student. Our approach can still obtain improvement with biased conformers, outperforming state-of-the-art methods and software (e.g.rdkit) generated 3D coordinates-based results. These phenomenon demonstrate the effectiveness of our distilling framework.



\begin{table}[!t]
\caption{The results of our method compared with other methods on the test set of MOLHIV dataset.}
\label{tab:hiv}
\centering
\setlength{\tabcolsep}{4.1mm}{
\begin{tabular}{cc}
\toprule
Method & AUC    \\
\midrule
DeeperGCN-FLAG\cite{kipf2016semi}  &  79.42 ± 1.20   \\
PNA \cite{kipf2016semi}  &  79.05 ± 1.32   \\
DGN  \cite{xu2018powerful} & 79.70 ± 0.97  \\
PHC-GNN\cite{ kipf2016semi} &  79.34 ± 1.16  \\
(baseline)Graphormer  \cite{ying2021transformers} & 80.51 ± 0.53  \\    
GraphMVP \cite{liu2021pre}   & 77.0   \\
GeomGCL\cite{li2022geomgcl} &  80.6 ± 0.009 \\
EGT  \cite{hussain2022global} & 80.60 ± 0.65  \\
GRPE-Standard  \cite{park2022grpe} & 81.39 ± 0.49  \\
GPS  \cite{rampavsek2022recipe} & 78.80 ± 0.0101  \\
 Ours(baseline+$\mathcal{L}_{m}$) & \textbf{81.4±0.11 }  \\
 Ours(baseline+$\overline{\mathcal{L}_{a}}$) &  79.0±0.01  \\
Ours(baseline+$\mathcal{L}_{m}$ +($\mathcal{W}_{search}*\overline{\mathcal{L}_{a}})$) & \textbf{81.47±0.03} \\
Ours(baseline+$\mathcal{L}_{m}$ +($\mathcal{W}_{coordinating}*\overline{\mathcal{L}_{a}})$) & \textbf{81.9±0.01} \\
\bottomrule
\end{tabular}}
\end{table}

\subsection{Ablation Study}
We investigated the impact of these components of our proposed method on the validation set of PCQM4Mv2 dataset, and results are shown in Table \ref{tab:pcqv2_abl}.

\begin{table}[!t]
\caption{The ablation study of our method on the validation set of the PCQM4Mv2 dataset. L denotes the conduction on the last layer and "all" denotes distilling with the features, which are not divided into global and local respectively. Our contributions
have a significant impact on performance.}
\label{tab:pcqv2_abl}
\centering
\setlength{\tabcolsep}{6.0mm}{
\begin{tabular}{cc}
\toprule
Method & MAE    \\
\midrule
Graphormer  \cite{ying2021transformers} & 0.0864 \\
baseline(reproduce+APE)   &  0.0845  \\
baseline+3D   &  0.040  \\
(L)baseline+all  &  0.0853  \\
(L)baseline+$\mathcal{L}_{m}$  &  0.0842  \\
(L)baseline+$\overline{\mathcal{L}_{a}}$ & 0.0874  \\
(L)baseline $+$ $\mathcal{L}_{m}$ $+$ $\mathcal{W}_{coordinating}$ $*$($\overline{\mathcal{L}_{a}}$)  &  0.0837  \\
baseline+$\mathcal{L}_{m}$  &  0.0822  \\
baseline+$\overline{\mathcal{L}_{a}}$ & 0.087  \\
baseline$+$ $\mathcal{L}_{m}$ $+$ $\mathcal{W}_{search}*$($\overline{\mathcal{L}_{a}}$)&  0.0818  \\
baseline$+$ $\mathcal{L}_{m}$ $+$ $\mathcal{W}_{coordinating}*$($\overline{\mathcal{L}_{a}}$)  &  0.0809  \\

\bottomrule
\end{tabular}}
\end{table}
\paragraph{The validation of absolute position encoding (APE)}

We train our graphormer with APE for 80 epochs as the baseline in Table \ref{tab:pcqv2_abl}. Compared to the original graphormer training with 300 epochs,  our baseline obtains a 1.52\% improvement in the second row.

\paragraph{The validation of dividing local atom and global molecular}
We observed that distilling the feature encoded with virtual token and atom token in the fourth row in Table \ref{tab:pcqv2_abl} generates a negative gain decreased by 0.94\% compared with baseline. The phenomenon indicates binding virtual token and atom tokens damages the performance of the distillation. 
\paragraph{The validation of global molecular}
Compared to the baseline model, the global molecular distillation $\mathcal{L}_{m}$  slightly increases the performance by distilling the virtual token embedding which is encoded with the structure information, reaching 0.36\% improvement in the fifth row in Table \ref{tab:pcqv2_abl}. 
\paragraph{The validation of layers}
In order to expand the effect of distilling molecular representation, we conduct the distillation on all layers, which generates 2.72\% improvement compared with conducting on the last layer. This result demonstrates that distilling on all layers superiors to distilling on the last layer.
\paragraph{The validation of local atoms}
In addition, we employ local atom loss $\mathcal{L}_{a}$ on all layers. We have observed one interesting phenomenon: when only leveraging the local atom information distillation, the network decreased by 3.43\% in sixth row. 
This phenomenon demonstrates that distilling local
atom information without correction will cause degradation problems.


\paragraph{The validation of search weight}

In the seventh row, the network further increases 3.2\% compared with baseline by distilling local atom and global molecular representation with coordinating weight searched manually denoted as $\mathcal{W}_{search}$. The result indicates that local atoms with coordinating weight can boost the performance of global molecular. But the optimal coordinating weight is hard to search manually due to the variable molecular size.
\paragraph{The validation of coordinating weight}
Finally, when we incorporate the global molecular and local atoms with coordinating weight $\mathcal{W}_{coordinating}$ to balance the grads according to the number of atoms, our method in the eleventh row achieves the best result with 0.809 MAE which achieves large improvement by 4.26\% compared with baseline. 
These results validate that our proposed coordination distillation with global molecular and local atom information is effective for molecular property prediction.

\subsection{Visionlization of effectiveness}

\begin{figure}[!t]
\centerline{\includegraphics[width=\linewidth]{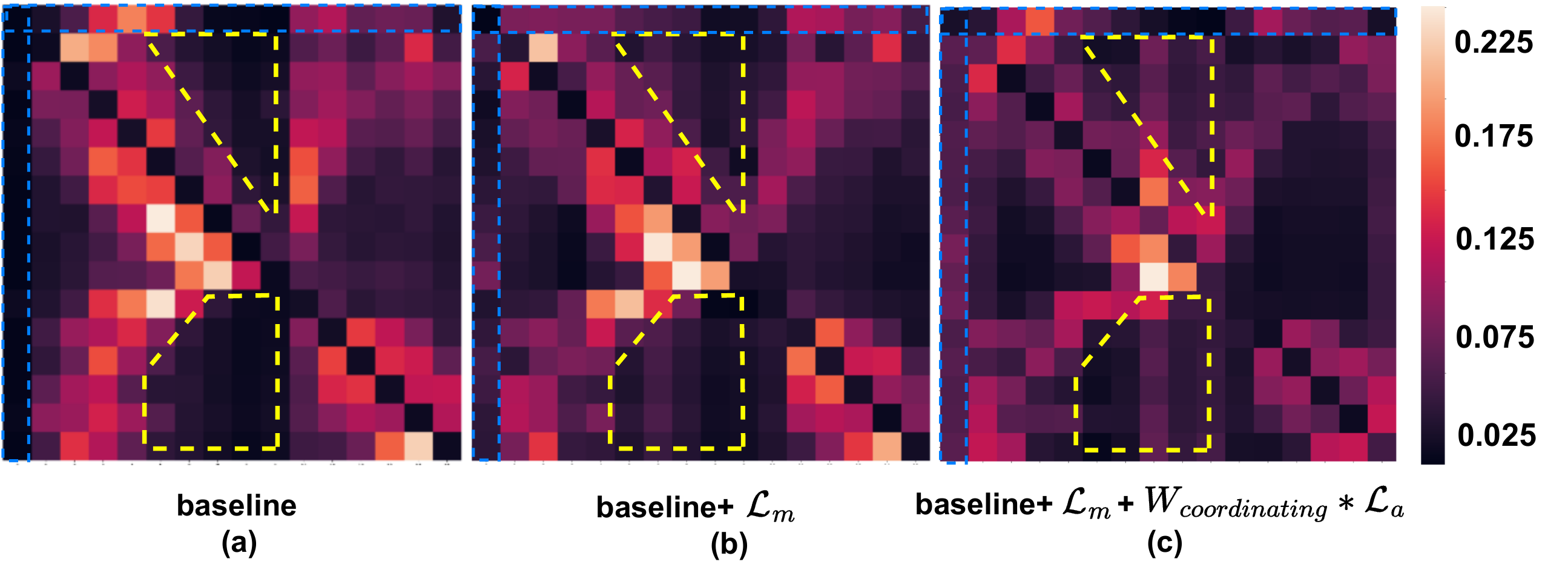}}
\caption{The visionlization of student model according to the attention weight. 
We compare the attention of atom tokens in yellow dashed fields and virtual token in the blue dashed boxes.} 
\label{fig:vision}
\end{figure}

\begin{figure}[!t]
\centerline{\includegraphics[width=\linewidth]{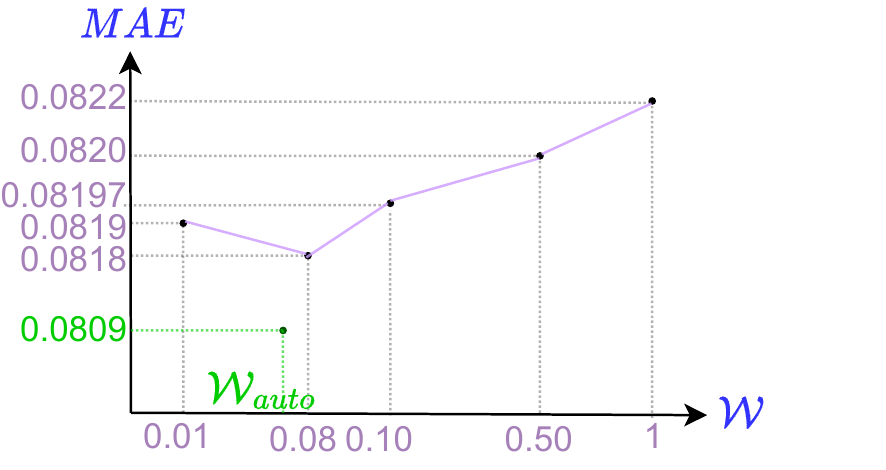}}
\caption{The comparison between coordinating weight and manually searched weights. The purple points are the results with manually searched weights, and the green point is the result with coordinating weight according to the number of atoms.}
\label{fig:zhexian}
\end{figure}

To understand and analyze the availability of our approach, we show the variation of attention weight in Fig. \ref{fig:vision}. 
We observed that the correlation between points is more prominent and notable compared with the baseline model. Compared with the baseline that all atoms only focus on the 2-hop nearest neighbors, both the global molecular token and local atom tokens could focus on almost all the atoms with CCMD. Thus, CCMD enriched the student model with more structured information.
This phenomenon promulgates the significance of distillation. 
To reveal the significance of coordinating weight, we certificate the validation of coordinating distilling strategy in Fig. \ref{fig:zhexian}, which shows that coordinating weight is superior to any manually searched weight. 
To analyze the behavior of the model after adding the distillation loss term during training, we present the learning curves of 2D supervised performance and distilling performance in Fig. \ref{fig:curve}. We can observe that the convergence of MAE is more stable compared with only 2D supervised loss.

\subsection{Effectiveness with Other Backbone}
To evaluate the effectiveness of our approach, we change the gragh transformer backbone of 2D branch into GIN \cite{xu2018powerful} architecture (L = 3, d = 768) on the PCQM4Mv2 dataset in Table \ref{tab:gin_abl}. 
The 3D model based on gragh transformer is a heavy model which has 12 blocks and 57.65 MB parameters, and the 2D model based on GIN is a light GNN model which has 12 blocks and 1.923 MB parameters. GIN and gragh transformer are very different in structure, making 2D-3D distilling much more challenging, while our method in the fourth row still increases MAE by 3.36\% compared with baseline in the first row.
What's more, the distillation with coordinating weight as $\frac{1}{N}$ in the fourth row increases 0.67\% compared with coordinating weight as $\frac{1}{{N}^{2}}$ in the third row.
This phenomenon indicates that coordinating weight with $\frac{1}{N}$ is enough to scale the gradients for the GNN network.
These results in Table \ref{tab:gin_abl} confirm the scalability and effectiveness of the distilling strategy.

\begin{table}[!t]
\caption{The results with GIN backbone on the validation set of PCQM4Mv2 dataset. 
}
\label{tab:gin_abl}
\centering
\setlength{\tabcolsep}{12.0mm}{
\begin{tabular}{cc}
\toprule
Method & MAE    \\
\midrule
baseline   &  0.122  \\
baseline+$\mathcal{L}_{m}$ &  0.1198  \\
baseline+$\mathcal{L}_{m}$+ $\frac{1}{N^{2}}*(\mathcal{L}_{a})$&  0.1187  \\
baseline+$\mathcal{L}_{m}$ + $\frac{1}{N}*\mathcal{L}_{a}$ &  \textbf{0.1179}  \\
\bottomrule
\end{tabular}}
\end{table}

\section{Conclusion}
In this work, we exclusively propose a novel CCMD framework for molecular property prediction.
In particular, we integrate global molecular token distillation and local atom information distillation with a novel coordinating weight to balance and scale the gradient magnitudes.
In addition, we establish an absolute position encoding to integrate adjacent edge information.
Experimental results on two popular molecular property prediction tasks demonstrate the superiority of our model.
In the future, we will further experiment on the datasets from other areas to demonstrate the generalization ability of our method and explore more superior distilling strategy.


\clearpage

\bigskip

\bibliographystyle{unsrt}
\bibliography{aaai23}

\begin{thebibliography}{10}

\bibitem{scarselli2008graph}
Franco Scarselli, Marco Gori, Ah~Chung Tsoi, Markus Hagenbuchner, and Gabriele
  Monfardini.
\newblock The graph neural network model.
\newblock {\em IEEE transactions on neural networks}, 20(1):61--80, 2008.

\bibitem{gilmer2017neural}
Justin Gilmer, Samuel~S Schoenholz, Patrick~F Riley, Oriol Vinyals, and
  George~E Dahl.
\newblock Neural message passing for quantum chemistry.
\newblock In {\em International conference on machine learning}, pages
  1263--1272. PMLR, 2017.

\bibitem{liu2019n}
Shengchao Liu, Mehmet~F Demirel, and Yingyu Liang.
\newblock N-gram graph: Simple unsupervised representation for graphs, with
  applications to molecules.
\newblock {\em Advances in neural information processing systems}, 32, 2019.

\bibitem{xu2021self}
Minghao Xu, Hang Wang, Bingbing Ni, Hongyu Guo, and Jian Tang.
\newblock Self-supervised graph-level representation learning with local and
  global structure.
\newblock In {\em International Conference on Machine Learning}, pages
  11548--11558. PMLR, 2021.

\bibitem{klicpera2020directional}
Johannes Klicpera, Janek Gro{\ss}, and Stephan G{\"u}nnemann.
\newblock Directional message passing for molecular graphs.
\newblock {\em arXiv preprint arXiv:2003.03123}, 2020.

\bibitem{qiao2020orbnet}
Zhuoran Qiao, Matthew Welborn, Animashree Anandkumar, Frederick~R Manby, and
  Thomas~F Miller~III.
\newblock c.
\newblock {\em The Journal of chemical physics}, 153(12):124111, 2020.

\bibitem{liu2021spherical}
Yi~Liu, Limei Wang, Meng Liu, Xuan Zhang, Bora Oztekin, and Shuiwang Ji.
\newblock Spherical message passing for 3d graph networks.
\newblock {\em arXiv preprint arXiv:2102.05013}, 2021.

\bibitem{toropov2005simplified}
Andrey~A Toropov, Alla~P Toropova, Dilya~V Mukhamedzhanoval, and Ivan Gutman.
\newblock Simplified molecular input line entry system (smiles) as an
  alternative for constructing quantitative structure-property relationships
  (qspr).
\newblock 2005.

\bibitem{hu2021ogb}
Weihua Hu, Matthias Fey, Hongyu Ren, Maho Nakata, Yuxiao Dong, and Jure
  Leskovec.
\newblock Ogb-lsc: A large-scale challenge for machine learning on graphs.
\newblock {\em arXiv preprint arXiv:2103.09430}, 2021.

\bibitem{ying2021transformers}
Chengxuan Ying, Tianle Cai, Shengjie Luo, Shuxin Zheng, Guolin Ke, Di~He,
  Yanming Shen, and Tie-Yan Liu.
\newblock Do transformers really perform badly for graph representation?
\newblock {\em Advances in Neural Information Processing Systems},
  34:28877--28888, 2021.

\bibitem{vaswani2017attention}
Ashish Vaswani, Noam Shazeer, Niki Parmar, Jakob Uszkoreit, Llion Jones,
  Aidan~N Gomez, {\L}ukasz Kaiser, and Illia Polosukhin.
\newblock Attention is all you need.
\newblock {\em Advances in neural information processing systems}, 30, 2017.

\bibitem{lu2019molecular}
Chengqiang Lu, Qi~Liu, Chao Wang, Zhenya Huang, Peize Lin, and Lixin He.
\newblock Molecular property prediction: A multilevel quantum interactions
  modeling perspective.
\newblock In {\em Proceedings of the AAAI Conference on Artificial
  Intelligence}, volume~33, pages 1052--1060, 2019.

\bibitem{liu2021pre}
Shengchao Liu, Hanchen Wang, Weiyang Liu, Joan Lasenby, Hongyu Guo, and Jian
  Tang.
\newblock Pre-training molecular graph representation with 3d geometry.
\newblock {\em arXiv preprint arXiv:2110.07728}, 2021.

\bibitem{li2022geomgcl}
Shuangli Li, Jingbo Zhou, Tong Xu, Dejing Dou, and Hui Xiong.
\newblock Geomgcl: Geometric graph contrastive learning for molecular property
  prediction.
\newblock In {\em Proceedings of the AAAI Conference on Artificial
  Intelligence}, volume~36, pages 4541--4549, 2022.

\bibitem{hu2019strategies}
Weihua Hu, Bowen Liu, Joseph Gomes, Marinka Zitnik, Percy Liang, Vijay Pande,
  and Jure Leskovec.
\newblock Strategies for pre-training graph neural networks.
\newblock {\em arXiv preprint arXiv:1905.12265}, 2019.

\bibitem{stark20223d}
Hannes St{\"a}rk, Dominique Beaini, Gabriele Corso, Prudencio Tossou, Christian
  Dallago, Stephan G{\"u}nnemann, and Pietro Li{\`o}.
\newblock 3d infomax improves gnns for molecular property prediction.
\newblock In {\em International Conference on Machine Learning}, pages
  20479--20502. PMLR, 2022.

\bibitem{jiao2019tinybert}
Xiaoqi Jiao, Yichun Yin, Lifeng Shang, Xin Jiang, Xiao Chen, Linlin Li, Fang
  Wang, and Qun Liu.
\newblock Tinybert: Distilling bert for natural language understanding.
\newblock {\em arXiv preprint arXiv:1909.10351}, 2019.

\bibitem{sun2019patient}
Siqi Sun, Yu~Cheng, Zhe Gan, and Jingjing Liu.
\newblock Patient knowledge distillation for bert model compression.
\newblock {\em arXiv preprint arXiv:1908.09355}, 2019.

\bibitem{wang2020minilm}
Wenhui Wang, Furu Wei, Li~Dong, Hangbo Bao, Nan Yang, and Ming Zhou.
\newblock Minilm: Deep self-attention distillation for task-agnostic
  compression of pre-trained transformers.
\newblock {\em Advances in Neural Information Processing Systems},
  33:5776--5788, 2020.

\bibitem{tian2019contrastive}
Yonglong Tian, Dilip Krishnan, and Phillip Isola.
\newblock Contrastive representation distillation.
\newblock {\em arXiv preprint arXiv:1910.10699}, 2019.

\bibitem{dong2020distilling}
Jin Dong, Marc-Antoine Rondeau, and William~L Hamilton.
\newblock Distilling structured knowledge for text-based relational reasoning.
\newblock In {\em Proceedings of the 2020 Conference on Empirical Methods in
  Natural Language Processing (EMNLP)}, pages 6782--6791, 2020.

\bibitem{zhu2021stepping}
Wenhao Zhu, Ziyao Li, Lingsheng Cai, and Guojie Song.
\newblock Stepping back to smiles transformers for fast molecular
  representation inference.
\newblock {\em arXiv preprint arXiv:2112.13305}, 2021.

\bibitem{buhmann2000radial}
Martin~D Buhmann.
\newblock Radial basis functions.
\newblock {\em Acta numerica}, 9:1--38, 2000.

\bibitem{hu2020open}
Weihua Hu, Matthias Fey, Marinka Zitnik, Yuxiao Dong, Hongyu Ren, Bowen Liu,
  Michele Catasta, and Jure Leskovec.
\newblock Open graph benchmark: Datasets for machine learning on graphs.
\newblock {\em Advances in neural information processing systems},
  33:22118--22133, 2020.

\bibitem{wu2018moleculenet}
Zhenqin Wu, Bharath Ramsundar, Evan~N Feinberg, Joseph Gomes, Caleb Geniesse,
  Aneesh~S Pappu, Karl Leswing, and Vijay Pande.
\newblock Moleculenet: a benchmark for molecular machine learning.
\newblock {\em Chemical science}, 9(2):513--530, 2018.

\bibitem{landrum2013rdkit}
Greg Landrum et~al.
\newblock Rdkit: A software suite for cheminformatics, computational chemistry,
  and predictive modeling.
\newblock {\em Greg Landrum}, 2013.

\bibitem{2017Automatic}
A.~Paszke, S.~Gross, S.~Chintala, G.~Chanan, E.~Yang, Z.~Devito, Z.~Lin,
  A.~Desmaison, L.~Antiga, and A.~Lerer.
\newblock Automatic differentiation in pytorch.
\newblock 2017.

\bibitem{xu2018powerful}
Keyulu Xu, Weihua Hu, Jure Leskovec, and Stefanie Jegelka.
\newblock How powerful are graph neural networks?
\newblock {\em arXiv preprint arXiv:1810.00826}, 2018.

\bibitem{kingma2014adam}
Diederik~P Kingma and Jimmy Ba.
\newblock Adam: A method for stochastic optimization.
\newblock {\em arXiv preprint arXiv:1412.6980}, 2014.

\bibitem{kipf2016semi}
Thomas~N Kipf and Max Welling.
\newblock Semi-supervised classification with graph convolutional networks.
\newblock {\em arXiv preprint arXiv:1609.02907}, 2016.

\bibitem{hussain2022global}
Md~Shamim Hussain, Mohammed~J Zaki, and Dharmashankar Subramanian.
\newblock Global self-attention as a replacement for graph convolution.
\newblock 2022.

\bibitem{park2022grpe}
Wonpyo Park, Woong-Gi Chang, Donggeon Lee, Juntae Kim, et~al.
\newblock Grpe: Relative positional encoding for graph transformer.
\newblock In {\em ICLR2022 Machine Learning for Drug Discovery}, 2022.

\bibitem{rampavsek2022recipe}
Ladislav Ramp{\'a}{\v{s}}ek, Mikhail Galkin, Vijay~Prakash Dwivedi, Anh~Tuan
  Luu, Guy Wolf, and Dominique Beaini.
\newblock Recipe for a general, powerful, scalable graph transformer.
\newblock {\em arXiv preprint arXiv:2205.12454}, 2022.

\end{thebibliography}


 \clearpage
\section{section 1}
\label{pro}
In this part, we will prove the relation of gradient on the virtual token and atom number ${N}$.

For transformer, to simplify the proof, $X_{i}^{l+1}$ in $l$ layer is formed as:
\begin{equation}
X_{i}^{l+1} =MHA(X^{l})=\sum_{j=0}^{N+1}({W}^{l}_{ji}*{X}^{l}_{j})
\end{equation}
The ${W}^{l}_{ji}$ is the attention weight, and ${W}^{l}_{ji}$ is formulated as:
\begin{equation}
{W}^{l}_{ji}= (\frac{e^{{X}^{l}_{i}*{X}^{l}_{j}}}{\sum_{j=0}^{N+1}(e^{{X}^{l}_{i}*{X}^{l}_{j})}})
\end{equation}
So,
\begin{equation}
\begin{aligned}
 {grad(X_{0}^{l})} &= \nabla_{X_{0}^{l}}{L} \\
&=\sum_{k=0}^{N+1}((\nabla_{X_{k}^{l+1}}{L})*(\nabla_{X_{0}^{l}}{X_{k}^{l+1}})) \\
&=\sum_{k=0}^{N+1}((\nabla_{X_{k}^{l+1}}{L})*(\nabla_{X_{0}^{l}}(\sum_{j=0}^{N+1}{W}_{jk}*{X}^{l}_{j})) \\
&=\sum_{k=0}^{N+1}((\nabla_{X_{k}^{l+1}}{L})*(\nabla_{X_{0}^{l}}((\sum_{j=1}^{N+1}{W}_{jk}*{X}^{l}_{j}) + {W}_{0k}*{X}^{l}_{0})) \\
&=\sum_{k=0}^{N+1}((\nabla_{X_{k}^{l+1}}{L})*((\sum_{j=1}^{N+1}\nabla_{X_{0}^{l}}({{W}_{jk})*{X}^{l}_{j}}) + {W}_{0k}) \\
&\propto(\sum_{k=0}^{N+1}(\nabla_{X_{k}^{l+1}}{L})*(\sum_{j=1}^{N+1}\nabla_{X_{0}^{l}}({{W}_{jk})*{X}^{l}_{j}}) + {O}({N}) \\
&\propto{O}(N)*((\sum_{j=1}^{N+1}\nabla_{{X}_{0}^{l}}{{W}_{jk}})+1)
\end{aligned}
\end{equation}
Now, we calculate ${\nabla_{{X}_{0}}{{W}_{jk}}}$ of the l-th layer, for convenience, ${a}$ and ${b}$ denote constants independent of the variable N and ${X}_{0}$:
\begin{equation}
\begin{aligned}
\nabla_{{X}_{0}}{{W}_{jk}}&=\nabla_{{X}_{0}}(\frac{e^{{X}_{j}*{X}_{k}}}{\sum_{j=0}^{N+1}e^{{X}_{j}*{X}_{k}}}) \\
&\propto(\nabla_{{X}_{0}}(\frac{a}{\sum_{j=0}^{N+1}e^{{X}_{j}*{X}_{k}}})) \\
&\propto(\nabla_{{X}_{0}}(\frac{a}{\sum_{j=1}^{N+1}e^{{X}_{j}*{X}_{k}}+e^{{X}_{0}*{X}_{k}}})) \\
&\propto(\nabla_{{X}_{0}}(\frac{a}{b+e^{{X}_{0}*{X}_{k}}})) \\
&\propto({O}(1))
\end{aligned}
\end{equation}


Finally, the virtual token gradient $grad(X_{0}^{l})$ in $l$ layer can be described in follow:
\begin{equation}
\begin{aligned}
 {grad(X_{0}^{l})} &\propto {{O}({N})} * \sum_{j=1}^{N+1}({O}(1)) \\
 &\propto{{O}({N}^{2})},
 \end{aligned}
\end{equation}
For a simple GNN like GIN:
\begin{equation}
\begin{aligned}
{X}^{l+1}_{i} &= {W}^{l+1}_{0i} * {X}^{l}_{0}
\end{aligned}
\end{equation}
So,
\begin{equation}
\begin{aligned}
 {grad(X_{0}^{l})} &= \nabla_{X_{0}^{l}}{L} \\
&=\sum_{k=0}^{N+1}((\nabla_{X_{k}^{l+1}}{L})*(\nabla_{X_{0}^{l}}{X_{k}^{l+1}})) \\
&=\sum_{k=0}^{N+1}((\nabla_{X_{k}^{l+1}}{L})*({W}^{l+1}_{0k})) \\
&\propto{O}(N)
\end{aligned}
\end{equation}
\end{document}